\documentclass[10pt,twocolumn,letterpaper,table,dvipsnames]{article}

\usepackage{cvpr}              
\usepackage{graphicx}
\usepackage{amsmath}
\usepackage{amssymb}
\usepackage{booktabs}
\usepackage{gensymb}
\usepackage{bm}
\usepackage{multirow}

\usepackage[dvipsnames]{xcolor}

\definecolor{cvprblue}{rgb}{0.21,0.49,0.74}
\usepackage[pagebackref,breaklinks,colorlinks,citecolor=cvprblue]{hyperref}
\definecolor{Gray}{gray}{0.6}
\definecolor{LightCyan}{rgb}{0.88,0.95,1}
\definecolor{blond}{rgb}{0.98, 0.94, 0.75}
\definecolor{light_gray}{gray}{0.95}

\def \ie {\emph{i.e.}}
\def \eg {\emph{e.g.}}
\def \etal {\emph{et al.}}

\newcommand{\tit}[1]{\smallbreak\noindent\textbf{#1.}}

\def\paperID{28} 
\def\confName{CVPR}
\def\confYear{2024}

\title{AIGeN: An Adversarial Approach for Instruction Generation in VLN}

\author{Niyati Rawal \quad Roberto Bigazzi \quad Lorenzo Baraldi \quad Rita Cucchiara \\
University of Modena and Reggio Emilia\\
{\tt\small \{name.surname\}@unimore.it}
}

\begin{document}
\maketitle
\begin{abstract}
In the last few years, the research interest in Vision-and-Language Navigation (VLN) has grown significantly. VLN is a challenging task that involves an agent following human instructions and navigating in a previously unknown environment to reach a specified goal. Recent work in literature focuses on different ways to augment the available datasets of instructions for improving navigation performance by exploiting synthetic training data. In this work, we propose AIGeN, a novel architecture inspired by Generative Adversarial Networks (GANs) that produces meaningful and well-formed synthetic instructions to improve navigation agents' performance. The model is composed of a Transformer decoder (GPT-2) and a Transformer encoder (BERT). During the training phase, the decoder generates sentences for a sequence of images describing the agent's path to a particular point while the encoder discriminates between real and fake instructions. Experimentally, we evaluate the quality of the generated instructions and perform extensive ablation studies. Additionally, we generate synthetic instructions for 217K trajectories using AIGeN on Habitat-Matterport 3D Dataset (HM3D) and show an improvement in the performance of an off-the-shelf VLN method. The validation analysis of our proposal is conducted on REVERIE and R2R and highlights the promising aspects of our proposal, achieving state-of-the-art performance.
\end{abstract}    
\section{Introduction}
\label{sec:intro}
In the last decade, we have witnessed remarkable results in Natural Language Processing, Computer Vision, and Robotics~\cite{bigazzi2024mapping}. More recently, increasing interest has been devoted to research at the intersection between these three domains~\cite{landi2022spot,poppi2023towards,bigazzi2023embodied}. In line with this trend, our work focuses on the Vision-and-Language Navigation (VLN) task. When performing VLN, an agent or a robot can perceive the 360\degree~view of the environment and is given human instructions such as ``\emph{Go to the living room and bring me the remote on the table}''. The agent has to follow the instructions and navigate an unknown environment to reach the specified goal and stop there.

\begin{figure*}[!t]
\centering
\includegraphics[width=.75\linewidth]{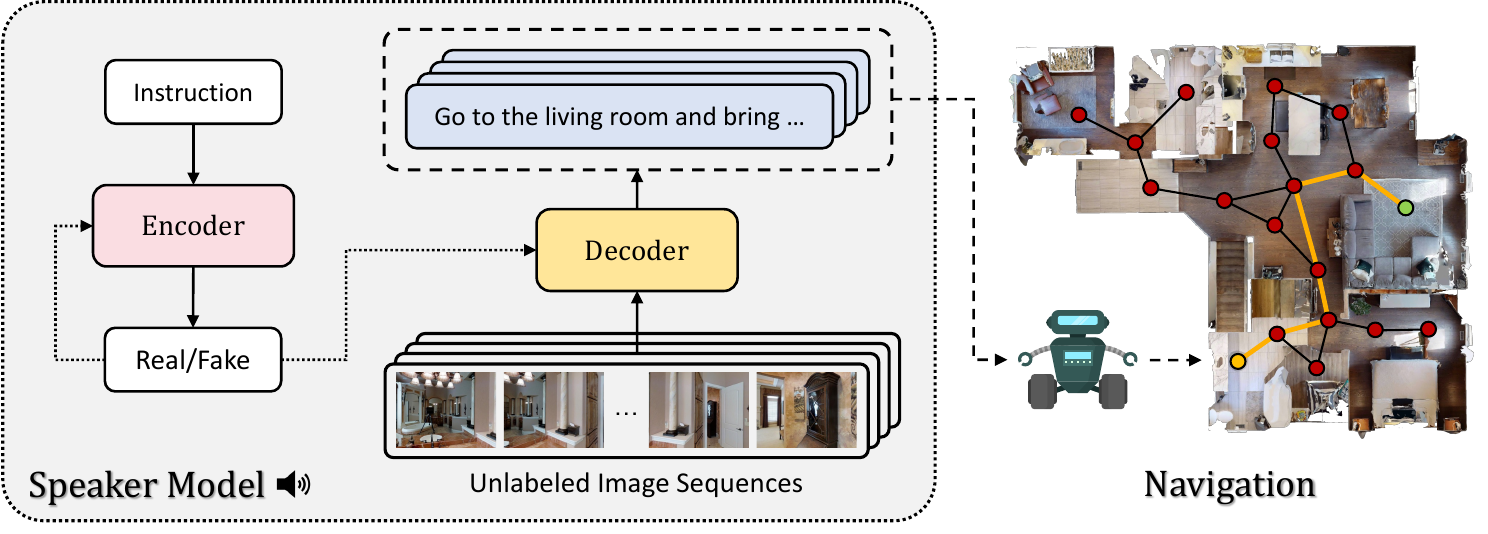}
\caption{We present AIGeN, a novel GAN-like model for generating instructions given a sequence of images. Synthetic instructions can be used as training data for a VLN model to improve its navigation performance.}
\label{fig:overview}
\vspace{-2mm}
\end{figure*}

However, human-generated instructions needed for training such architectures are costly and time-consuming to obtain. The resulting lack of annotated training data is one of the key factors in making VLN a challenging task. Recent work~\cite{fried2018speaker,guhur2021airbert,wang2021less,chen2022learning} has focused on generating instructions at a lower cost by employing methods for synthetic instruction generation. For example, Guhur~\etal~\cite{guhur2021airbert} and Chen~\etal~\cite{chen2022learning} showed that generating synthetic instructions and augmenting the data, improves the navigation performance of the agent.
Nevertheless, Guhur~\etal~\cite{guhur2021airbert} used image-caption pairs from the web on a prohibitive total number of $140$K environments. Instead, Chen~\etal~\cite{chen2022learning} generated synthetic instructions using trajectories sampled on HM3D dataset~\cite{ramakrishnan2021habitat} which is composed of 900 environments; our approach aims at improving the generation quality of such instructions. 

As can be seen in Fig.~\ref{fig:overview}, 
we propose AIGeN, a novel computational model that can generate synthetic instructions starting from unlabeled navigation paths in an environment. Our proposed model combines a multimodal Generative Pre-Trained Transformer (GPT)~\cite{brown2020language} and Bidirectional Encoder Representations from Transformers (BERT)~\cite{devlin2018bert} in an adversarial manner to generate high-quality instructions. 
In particular, the model consists of a Transformer decoder (GPT-2) that generates sentences describing the agent's path, using a sequence of images from the environment and the associated object detections extracted using Mask2Former~\cite{cheng2022masked}. The BERT-like encoder, instead, serves as a discriminator and is trained to distinguish between real and fake instructions. 

Using our approach to augment the training data of REVERIE and R2R datasets, we show that our AIGeN-generated instructions help to improve the results of a VLN model achieving state-of-the-art performance.
Additionally, we validate the quality of the proposed method by evaluating the generated instructions using image description metrics~\cite{stefanini2022show} and comparing the downstream navigation performance of different configurations of our model, showing that producing well-formed synthetic instructions is beneficial for the training of a navigation agent. 
\section{Related Work}
\label{sec:relatedWork}

\tit{Vision-and-Language Navigation}
VLN is a task where an agent needs to learn to navigate in a previously unknown environment by following step-by-step natural language instructions~\cite{anderson2018vision,gu2022vision}. Thus, the VLN task requires the agent to understand visual and textual cues and their correlation and plan its movements accordingly. This makes VLN a challenging task, which has attracted remarkable research efforts. While this work focuses on VLN in indoor environment, there are also works that emphasize VLN in outdoor environments~\cite{chen2019touchdown,mehta2020retouchdown,zhu2020multimodal}
Early work proposed a sequence-to-sequence long short-term memory for action inference~\cite{anderson2018vision} and adopted a panoramic action space with a module to generate synthetic instructions~\cite{fried2018speaker}. Ma~\etal~\cite{ma2019self} proposed a self-monitoring agent, and a method combining progress inference with learned heuristic~\cite{ma2019regretful}. Landi~\etal~\cite{landi2019embodied} used dynamic convolution filters. RCM~\cite{wang2019reinforced}, instead, used reinforcement learning to improve cross-modal matching.
Recently, Transformer-based models~\cite{vaswani2017attention} have been gaining popularity among VLN methods. For example, PRESS~\cite{li2019robust} applied pretrained BERT to encode instructions. VLN$\circlearrowright$BERT~\cite{hong2021vln} implemented a recurrent BERT to model time dependencies. SIA~\cite{lin2021scene} and PTA~\cite{landi2021multimodal} used Transformers for multimodal fusion, and HAMT~\cite{chen2021history} directly used Transformers to exploit episode history. Among graph-based methods, Hong~\etal~\cite{hong2020language} implemented graphs to model relations between the scene, objects, and instructions, while DUET~\cite{chen2022think} and KERM~\cite{li2023kerm} used topological maps with a dual-scale graph Transformer to encode long-term action planning and fine-grained cross-modal understanding. Additionally, AZHP~\cite{gao2023adaptive} proposed a hierarchical navigation process instead of a single-step planning scheme.

\begin{figure*}[!t]
\centering
\includegraphics[width=.85\linewidth]{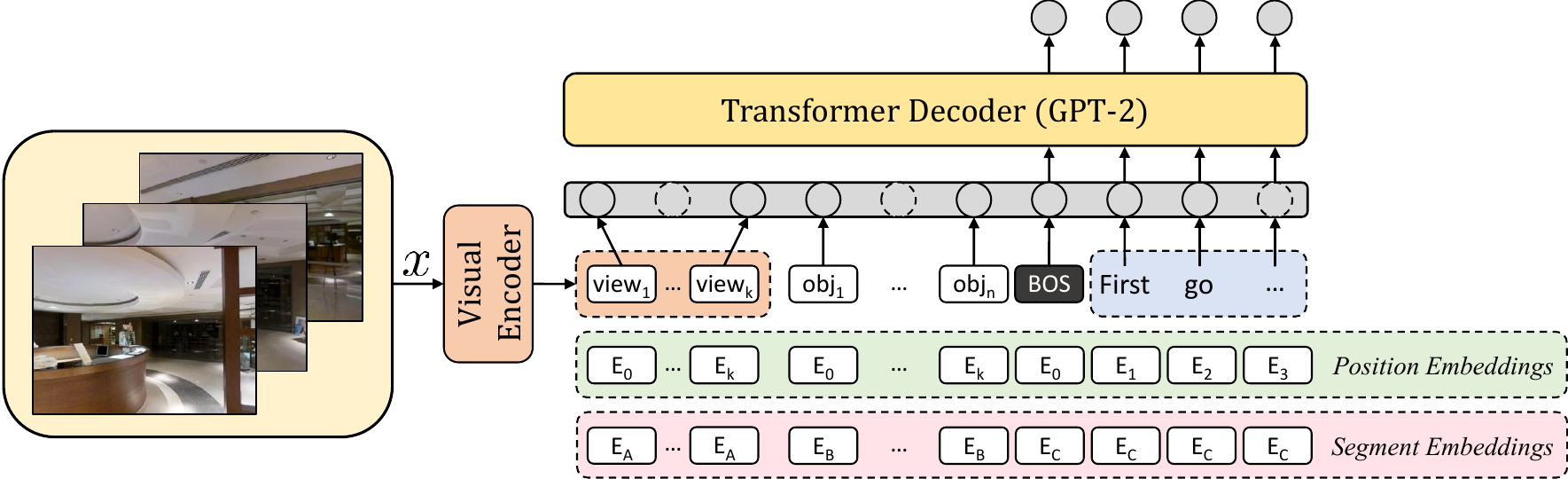}
\caption{Architecture of the instruction generator model that generates VLN instructions. The model is fed with a sequence of images and object names, and it is prompted to generate instructions using a \texttt{BOS} token. The instructions are generated token by token until the \texttt{EOS} token.
} 
\vspace{-2mm}
\label{fig:generator}
\end{figure*}

\tit{Instruction Generation}
A different direction of research on VLN is followed by studies that present methods for augmenting available datasets.
Besides the already mentioned Fried~\etal~\cite{fried2018speaker}, Majumdar~\etal~\cite{majumdar2020improving} proposed a Transformer model that transfers visual grounding learned from image-text pairs from the web to VLN.
Other research work samples random trajectories and generates instructions using online rental marketplaces~\cite{guhur2021airbert} and a large-scale dataset of indoor environments~\cite{chen2022learning,kamath2023new}.
Our approach improves the generation of synthetic instructions by aligning the visual input features with the generated text using the same number of trajectories as Chen~\etal~\cite{chen2022learning}.

The most common Transformer-like networks used in Natural Language Processing (NLP) applications are BERT~\cite{devlin2018bert}, GPT~\cite{brown2020language}, BART~\cite{lewis2019bart} and Text-to-Text Transfer Transformer (T5)~\cite{raffel2019exploring}. While BERT consists of only the encoder part of the Transformer, GPT consists of its decoder part. BERT is beneficial for tasks like Visual Question Answering (VQA), \eg~\cite{lu2019vilbert,tan2019lxmert}. On the other hand, GPT is useful for text generation or summarization~\cite{brown2020language}.
Moreover, Visual Language Models (VLMs)~\cite{alayrac2022flamingo} have been used to generate text, given images, and textual words.
Chen~\etal~\cite{chen2022learning} proposed to use a GPT-2 decoder that acts like a speaker model to generate instructions for VLN. 
In this study, we use a GPT-2 decoder that generates language instructions, while BERT encoder discriminates whether the instructions correspond to the given sequence of images as real or fake.

\tit{Generative Adversarial Networks}
While GANs have been applied effectively for generating images~\cite{goodfellow2014generative}, they have been seldom applied to generate instructions that consist of discrete tokens as it is difficult to backpropagate through discrete tokens. Previous works rely on REINFORCE~\cite{dai2017towards} or using Gumbel-Softmax~\cite{shetty2017speaking} to allow for backpropagation. Recently, Transformer networks were used in an adversarial manner, however, these approaches were used for the generation of images~\cite{jiang2021transgan, hudson2021generative}.

\section{Proposed Method}
\label{sec:approach}
\begin{figure*}[!t]
\centering
\includegraphics[width=.85\linewidth]{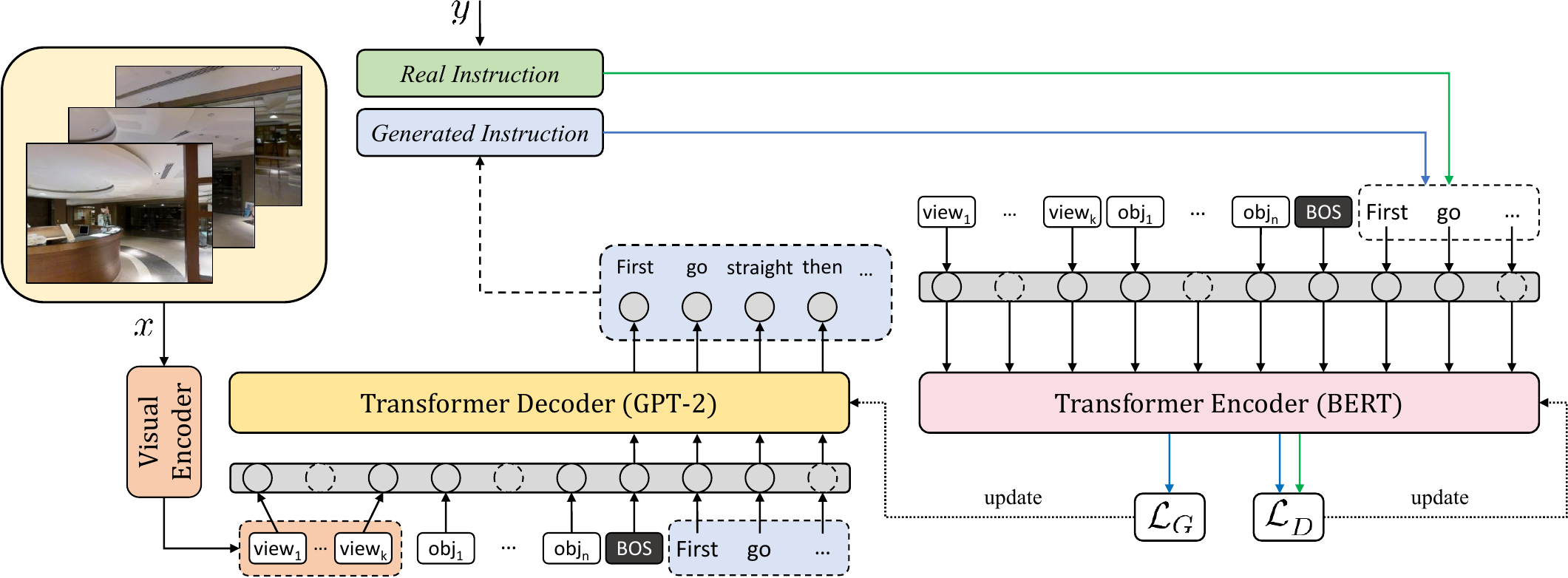}
\caption{Schema of the proposed generative-adversarial framework for synthetic instruction generation. The GPT-2 decoder acts as a generator while the BERT encoder acts as a discriminator. Both models are trained simultaneously. The generator generates fake instructions token by token until it reaches the \texttt{EOS} token. The discriminator has to detect whether the instructions corresponding to a given sequence of images is real (ground-truth) or fake (generated).}
\vspace{-2mm}
\label{fig:model}
\end{figure*}

AIGeN is used to describe an agent's trajectory in natural language and is composed of an instruction generator (visually depicted in Fig.~\ref{fig:generator}) and an instruction discriminator. Both models are trained simultaneously using an adversarial approach, which aims at improving the performance of the generator and the quality of the generated instructions. The overall model is shown in Fig.~\ref{fig:model}.
To the best of our knowledge, this is the first approach that combines Transformer networks and a GAN-like training procedure for generating synthetic navigation instructions.

\subsection{Instruction Generator}
\label{subsec:generator}
The generator is a language model in charge of describing the actions needed to reach the target location. The proposed model exploits a pretrained language model which is finetuned conditioning on visual inputs to achieve multimodal capabilities similar to Alayrac~\etal~\cite{alayrac2022flamingo}.
However, instead of just using GPT-2 decoder for instruction generation, we use it in combination with BERT in a GAN-like manner to generate better-quality instructions. 

The overall approach takes the images from the environment the agent has to traverse and feeds them into a pretrained ResNet-152~\cite{he2016deep}
encoder to obtain image features.
Next, when generating instructions on Habitat-Matterport 3D (HM3D) environments, all the objects in the given sequence of images are extracted using Mask2Former~\cite{cheng2022masked} trained on ADE20K to enrich the visual features with object names. 
This input is fed into the decoder along with the first \texttt{BOS} token using the concatenation of visual features and object names as a prompt for the language model.
The GPT-2 decoder is trained to predict the subsequent language tokens to generate a complete language instruction that describes the actions of the agent until the last image. 
Differently from~\cite{alayrac2022flamingo}, 
AIGeN flattens the tokens from both images and text before feeding them into the GPT-2 decoder. Furthermore, to effectively segregate the visual information from the textual information, position and segment embeddings are used in addition to the tokens, 
as shown in Fig.~\ref{fig:generator}.

The result of the generation is defined starting from the output of the decoder following the \texttt{BOS}, autoregressively generating instructions token by token and
allowing backpropagation through sampled tokens using the Gumbel-Softmax~\cite{shetty2017speaking} trick. 

\subsection{Instruction Discriminator}
\label{subsec:discriminator}
The inputs of the BERT encoder are image features, tokens of the names of objects in the images, and an instruction. Similarly to the inputs for GPT-2 decoder, the images are first fed into a pretrained ResNet-152~\cite{he2016deep} visual encoder to obtain image features.
Then, a fully connected layer followed by a sigmoid function is used to process the results from BERT output. The output is simply real or fake; real if the given instruction matches the given sequence of images and fake otherwise. Finally, binary cross-entropy loss is used to minimize the error between the generated output and the actual output (real or fake). A well-trained discriminator should classify ground-truth instructions as real and the generated instructions as fake.
Like the decoder, BERT input employs position embeddings and segment embeddings in addition to token embeddings.

\subsection{Adversarial Training}
Our method, shown in Fig.~\ref{fig:model}, follows an adversarial training approach, where
the generator $G$ is trained to fool the discriminator $D$, while the discriminator is taught to distinguish between real and fake instructions.
\noindent The generator is trained to generate instructions as close to the ground-truth instructions as possible by minimizing the cross-entropy loss between the generated instructions and the ground-truth instructions. The generator loss is defined as: 
\begin{equation}
\mathcal{L}_G=-\log(D(I_{G},\bm{x})),
\end{equation}
where $I_{G} \in G(\bm{x})$ is the generated instruction and $\bm{x}$ is the sequence of images belonging to the trajectory.

\noindent The discriminator has to discriminate between the ground-truth instructions as real instructions and the generated instructions as fake instructions. Consequently, the discriminator loss is:
\begin{equation}
\label{eq:d_loss}
    \mathcal{L}_D=-\log(1-D(I_{G},\bm{x})) -\log(D(I_{R},\bm{x})), \\
\end{equation} 
where $I_{R} \in R(\bm{x})$ is the ground-truth instruction. The training is performed simultaneously both on the generator and the discriminator. 
\section{Experiments}
\label{sec:experiments} 

\subsection{Experimental Setup}
We present experimental results using the model proposed in Section~\ref{sec:approach} employing two VLN datasets as benchmark,  REVERIE~\cite{qi2020reverie} and R2R~\cite{anderson2018vision}. Both datasets consist of navigation sequences composed of 360$\degree$ images that are collected on the nodes of navigation graphs on Matterport3D environments~\cite{Matterport3D} and each navigation sequence is associated with three ground-truth textual instructions.
The main difference between the two datasets is that REVERIE instructions include interactions with specific target objects that the model is required to identify, whereas R2R instructions only specify the trajectory to be followed.
In this study, we only consider the frontal view of the 360$\degree$ images, with a field of view of 60$\degree$.
The training of AIGeN uses a learning rate of 0.0002, batch size of 6, and Adam~\cite{kingma2014adam} as optimizer. For the GPT-2 decoder, unless otherwise stated we use a medium-sized version with $L=12$, $d=768$, $H=12$, where $L$ is the number of layers, $d$ is the model dimensionality and $H$ is the number of attention heads.  The visual features used by the model are extracted using ResNet-152.
For the BERT encoder, we use a hidden size of 768, 12 layers, and an intermediate size of 3072. Overall, AIGeN has $289$M model parameters and is trained for $\approx36$ hours on a single NVIDIA RTX6000 GPU.

\begin{table*}[t]
\centering
\setlength{\tabcolsep}{0.9em}
\resizebox{0.63\linewidth}{!}{
\begin{tabular}{l | cccc | cc}
\toprule
\multicolumn{1}{c}{} & \multicolumn{6}{c}{\textbf{Val Unseen}} \\
\addlinespace[4pt]
& TL & SPL$\uparrow$ & SR$\uparrow$ & OSR$\uparrow$ & RGS$\uparrow$ & RGSPL$\uparrow$ \\
\midrule
Seq2Seq~\cite{anderson2018vision} & 11.1 & 2.8 & 4.2 & 8.1 & 2.2 & 1.6 \\
\rowcolor{light_gray}
SMNA~\cite{ma2019self} & 9.1 & 6.4 & 8.2 & 11.3 & 4.5 & 3.6 \\
RCM~\cite{wang2019reinforced} & 12.0 & 7.0 & 9.3 & 14.2 & 4.9 & 3.9 \\
\rowcolor{light_gray}
FAST-MATTN~\cite{qi2020reverie} & 45.3 & 7.2 & 14.4 & 28.2 & 7.8 & 4.7 \\
SIA~\cite{lin2021scene} & 41.5 & 16.3 & 31.5 & 44.7 & 22.4 & 11.6 \\
\rowcolor{light_gray}
Airbert~\cite{guhur2021airbert} & 18.7 & 21.9 & 27.9 & 34.5 & 18.2 & 14.2 \\
ProbES~\cite{liang2022visual} & 18.0 & 22.8 & 27.6 & 33.2 & 16.8 & 13.9 \\
\rowcolor{light_gray}
VLN$\circlearrowright$BERT~\cite{hong2021vln} & 16.8 & 24.9 & 30.7 & 35.0 & 18.8 & 15.3 \\
HAMT~\cite{chen2021history} & 14.1 & 30.2 & 33.0 & 36.8 & 18.9 & 17.3 \\
\rowcolor{light_gray}
DUET~\cite{chen2022think} & 22.1 & 33.7 & 47.0 & 51.1 & 32.2 & 23.0 \\
KERM~\cite{li2023kerm} & 21.9 & 35.4 & 50.4 & 55.2 & 34.5 & 24.5 \\
\rowcolor{light_gray}
AZHP~\cite{gao2023adaptive} & 22.3 & 36.6 & 48.3 & 53.6 & 34.0 & 25.8 \\
HM3D-AutoVLN\textsuperscript{*}~\cite{chen2022learning} & 24.3 & 39.4 & 54.0 & \textbf{60.0} & 34.6 & 25.2 \\
\rowcolor{light_gray}
\textbf{DUET\ +\ AIGeN} & 19.5 & \textbf{41.9} & \textbf{54.4} & 57.7 & \textbf{35.1} & \textbf{26.9} \\
\bottomrule
\end{tabular}
}
\vspace{-2mm}
\caption{VLN metrics for our approach and competitors on REVERIE Val Unseen. \textsuperscript{*} denotes finetuning on our computing architecture.
} 
\label{tab:reverie_val_unseen}
\end{table*}

\begin{table*}[t]
\centering
\setlength{\tabcolsep}{.9em}
\resizebox{0.7\linewidth}{!}{
\begin{tabular}{l | cccc | cccc}
\toprule
\multicolumn{1}{c}{} & \multicolumn{4}{c}{\textbf{Val Unseen}} & \multicolumn{4}{c}{\textbf{Test Unseen}} \\
\addlinespace[4pt]
& TL & SPL$\uparrow$ & SR$\uparrow$ & NE$\downarrow$ & TL & SPL$\uparrow$ & SR$\uparrow$ & NE$\downarrow$ \\
\midrule
Seq2Seq~\cite{anderson2018vision} & 8.39 & - & 22 & 7.81 & 8.13 & 18 & 20 & 7.85 \\
\rowcolor{light_gray}
PRESS~\cite{li2019robust} & 10.36 & 45 & 49 & 5.28 & 10.77 & 45 & 49 & 5.49  \\
SSM~\cite{wang2021structured} & 20.70 & 45 & 62 & 4.32 & 20.39 & 46 & 61 & 4.57 \\ 
\rowcolor{light_gray}
EnvDrop~\cite{tan2019learning} & 10.70 & 48 & 52 & 5.22 & 11.66 & 47 & 51 & 5.23 \\
PREVALENT~\cite{hao2020towards} & 10.19 & 53 & 58 & 4.71 & 10.51 & 51 & 54 & 5.30 \\
\rowcolor{light_gray}
RelGraph~\cite{hong2020language} & 9.99 & 53 & 57 & 4.73 & 10.29 & 52 & 55 & 4.75 \\
ProbES~\cite{liang2022visual} & 11.58 & 55 & 61 & 4.03 & 12.43 & 56 & 62 & 4.20 \\
\rowcolor{light_gray}
Airbert~\cite{guhur2021airbert} & 11.78 & 56 & 62 & 4.01 & 12.41 & 57 & 62 & 4.13 \\
VLN$\circlearrowright$BERT~\cite{hong2021vln} & 12.01 & 57 & 63 & 3.93 & 12.35 & 57 & 63 & 4.09 \\
\rowcolor{light_gray}
MARVAL~\cite{kamath2023new} & 10.15 & 61 & 65 & 4.06 & 10.22 &  58 & 62 & 4.18 \\
DUET~\cite{chen2022think} & 13.94 & 60 & 72 & 3.31 & 14.73 & 59 & 69 & 3.65 \\
\rowcolor{light_gray}
KERM~\cite{li2023kerm} & 13.54 & 60 & 72 & 3.22 & 14.74 & 59 & 70 & 3.61 \\
HAMT~\cite{chen2021history} & 11.46 & 61 & 66 & \textbf{2.29} & 12.27 & 60 & 65 & 3.93 \\
\rowcolor{light_gray}
AZHP~\cite{gao2023adaptive} & 14.05 & 61 & 72 & 3.15 & 14.95 & 60 & \textbf{71} & 3.52 \\
\textbf{DUET\ +\ AIGeN} & 13.72 & \textbf{63} & \textbf{73} & 2.92 & 14.20 & \textbf{61} & \textbf{71} & \textbf{3.33}\\
\bottomrule
\end{tabular}
}
\vspace{-2mm}
\caption{VLN experiments on the Val Unseen split of R2R dataset.} 
\label{tab:r2r_val_unseen}
\end{table*}
We evaluate the navigation performance using the following navigation metrics:
trajectory length in meters (TL); success rate (SR), \ie~the fraction of episodes where the agent can reach the goal position within $3$ meters; oracle success rate (OSR), that is the success rate using an oracle stop policy; success rate weighted by path length (SPL); and navigation error (NE). The object grounding ability of the agents on REVERIE dataset is evaluated using remote grounding success (RGS) which is the ratio of successfully followed instructions, and RGS weighted by path length (RGSPL).

Instead, to quantitatively evaluate our model on text generation, we use metrics commonly used for image description~\cite{stefanini2022show}, namely BLEU~\cite{papineni2002bleu}, METEOR~\cite{banerjee2005meteor}, ROUGE~\cite{lin2004rouge},  CIDEr~\cite{vedantam2015cider}, and SPICE~\cite{spice2016}\footnote{We compute image description metrics using the code provided at the following link: \url{https://github.com/tylin/coco-caption}}. All these metrics are obtained by comparing the predicted instruction with the ground-truth instruction in terms of their \emph{n}-grams (where an \emph{n}-gram is a sequence of \emph{n} consecutive words). While all these metrics are commonly used for evaluating cross-modal description, only CIDEr and SPICE have been specifically designed for this task. The others (BLEU, METEOR, and ROUGE) have indeed been proposed for evaluating translation and summarization.
According to recent literature, CIDEr showcases the best alignment with human judgment.

\subsection{Experimental Results on VLN}
The navigation experiments on VLN are performed considering an off-the-shelf state-of-the-art VLN method finetuned with the instructions generated by our approach.
We adopt DUET~\cite{chen2022think}, which is based on a dual-scale graph Transformer to perform long-term action planning and fine-grained cross-modal understanding by exploiting topological maps that are built during the episode.
DUET has
a two-step training as described by Chen~\etal~\cite{chen2022think}, that is composed of a pretraining phase on four auxiliary tasks and the finetuning on the VLN task. We perform the pretraining phase following~\cite{chen2022learning}, and then we finetune the pre-trained model augmenting REVERIE and R2R training splits with our synthetic instructions. AIGeN-generated instructions are produced using $217$K randomly sampled trajectories on Habitat-Matterport 3D (HM3D)~\cite{ramakrishnan2021habitat} dataset, a large-scale dataset of indoor photorealistic environments. We use the same paths used and released by Chen~\etal~\cite{chen2022learning}, and we generate one synthetic instruction for each sequence of observations.
The navigation model is pre-trained and finetuned for a total of $\approx32$ hours, $12$ and $20$ respectively, on a single NVIDIA RTX6000 GPU, and is compared with current state-of-the-art VLN models on REVERIE and R2R.

\begin{table*}[!t]
\centering
\resizebox{.94\linewidth}{!}{
\setlength{\tabcolsep}{0.6em}
\begin{tabular}{p{29mm} | ccccc | ccccc}
\toprule
\multicolumn{1}{c}{} & \multicolumn{5}{c}{\textbf{Val Seen}} & \multicolumn{5}{c}{\textbf{Val Unseen}} \\
\addlinespace[4pt]
Model & BLEU-1 & METEOR & ROUGE & CIDEr & SPICE & BLEU-1 & METEOR & ROUGE & CIDEr & SPICE \\
\midrule
AIGeN (Medium) & \textbf{0.487} & 0.222 & 0.457 & 0.869 & 0.318 & 0.412 & 0.166 & 0.378 & 0.461 & 0.213 \\ 
\rowcolor{light_gray}
AIGeN w/o detect. & {0.409} & {0.188} & {0.416} & {0.358} & {0.196} & {0.357} & {0.141} & {0.347} & {0.132} & {0.123} \\
\textbf{AIGeN} & 0.484 & \textbf{0.228} & \textbf{0.465} & \textbf{0.890} & \textbf{0.329} & \textbf{0.421} & \textbf{0.179} & \textbf{0.393} & \textbf{0.486} & \textbf{0.228} \\
\bottomrule
\end{tabular}
}
\vspace{-2mm}
\caption{Image description experiments using our model with different size configurations and without using the detections on REVERIE.
}
\label{table:description_ablation}
\end{table*}

\begin{table*}[!t]
\centering
\setlength{\tabcolsep}{0.9em}
\resizebox{0.61\linewidth}{!}{
\begin{tabular}{l | cccc | cc}
\toprule
\multicolumn{1}{c}{} & \multicolumn{6}{c}{\textbf{Val Unseen}} \\
\addlinespace[4pt]
& TL & SPL$\uparrow$ & SR$\uparrow$ & OSR$\uparrow$ & RGS$\uparrow$ & RGSPL$\uparrow$ \\
\midrule
AIGeN (Medium) & 20.0 & 41.2 & 54.3 & \textbf{60.1} & \textbf{35.1} & 26.5 \\
\rowcolor{light_gray}
AIGeN w/o detect. & 23.4 & 38.9 & 52.5 & 59.4 & 34.6 & 25.7 \\
\textbf{AIGeN} & 19.6 & \textbf{41.9} & \textbf{54.4} & 57.7 & \textbf{35.1} & \textbf{26.9} \\
\bottomrule
\end{tabular}
}
\vspace{-2mm}
\caption{Experimental comparison of VLN performance of different configurations of our model on REVERIE dataset.
}
\vspace{-1mm}
\label{table:navigation_ablation}
\end{table*}

\begin{table*}[!t]
\centering
\setlength{\tabcolsep}{0.9em}
\resizebox{0.64\linewidth}{!}{
\begin{tabular}{l | cccccc}
\toprule
\addlinespace[4pt]
& \%Novel & Unigrams & Bigrams & Div-1 & Div-2 \\
\midrule
Ground truth & - & 3675 & 21551 & 0.019 & 0.113 \\
\rowcolor{light_gray}
AIGeN w/o adv. training & 22.7\% & 2970 & 15928 & 0.016 & 0.086 \\
\textbf{AIGeN} & 100.0\% & 14783 & 43013 & 0.072 & 0.210 \\
\bottomrule
\end{tabular}
}
\vspace{-1mm}
\caption{Evaluation of the diversity of synthetic instructions using AIGeN before and after the GAN fine-tuning compared with the ground truth instructions on REVERIE training split.
}
\label{tab:diversity}
\end{table*}

The navigation results on REVERIE in Table~\ref{tab:reverie_val_unseen} show that our approach achieves state-of-the-art performance on SPL and both object grounding metrics while remaining competitive on the other metrics. In particular, our instruction set gives a boost in generating effective trajectories towards the goal, which is reflected by SPL and RGSPL. For example, our approach 
shows an improvement of $8.2$ and $3.9$ on SPL and RGSPL with respect to the baseline DUET.
For fairness of comparison, we re-trained HM3D-AutoVLN on our computing infrastructure following~\cite{chen2022learning} (denoted as HM3D-AutoVLN\textsuperscript{*}). The comparison between HM3D-AutoVLN\textsuperscript{*} and our approach highlights the effectiveness of AIGeN-generated instruction since the difference between the two methods on VLN is defined by the quality of the synthetic instructions exploited to finetuned DUET. When running this comparison, our approach outperforms HM3D-AutoVLN* on the main navigation metrics, and in particular, on success rate weighted by path length (SPL) shows an improvement of $2.5$.

Furthermore, analyzing the experiments on R2R dataset in Tab.~\ref{tab:r2r_val_unseen}, our method provides state-of-the-art results on SPL and SR for both validation and test unseen splits, proving that our approach can generate synthetic instructions that are beneficial for multiple VLN datasets. 
Comparing our method with respect to the baseline DUET, AIGeN-generated instructions allow an improvement in SPL of $3.0$ and $2.0$ respectively on the validation ``unseen'' and test ``unseen'' splits.
Overall, these results support our claim that well-formed synthetic instructions help the agent to learn better navigation and object localization in a VLN setting.

\subsection{Ablation Study} 
We validate the components of our approach by comparing different configurations of AIGeN on synthetic instructions generation in Table~\ref{table:description_ablation} and on navigation in Table~\ref{table:navigation_ablation}.

Starting from Table~\ref{table:description_ablation}, the first row shows the performance of the generator that is trained from scratch. 
As can be seen, training the instruction generation model from scratch provides a maximum CIDEr of 0.869 on the ``seen'' split, and 0.378 on the ``unseen'' split, without reaching the results of the overall method.
Following, we ablate the proposed model in terms of input modalities analyzing the contribution given by the object detections, and comparing it with a model that is trained using only visual features and textual instructions. All the metrics related to AIGeN without detections, especially CIDEr and SPICE, are considerably lower than the metric values computed for AIGeN. This result confirms the importance of employing object detections as input features. We speculate that when there are no object words provided, the model is not able to identify which object in the scene it has to attend to. 
Therefore, the target objects in the generated sentences are often different from those in the ground-truth instructions even when the landmark is correctly recognized.

Moving on to the navigation experiments in Table~\ref{table:navigation_ablation}, we compare the performance of the VLN model on REVERIE dataset augmenting the training data with the instructions generated by different configurations of our approach. In this case, AIGeN surpasses its counterparts by an important margin on all the metrics. The improvement on the SPL over the Medium configuration that is trained from scratch is $0.7$, and it becomes $3.0$ when considering AIGeN w/o detections.
The ablation study validates the effectiveness of the components of the model in both text generation and downstream vision-and-language navigation.

\begin{figure*}[!t]
    \centering
    \small
    \resizebox{.99\linewidth}{!}{
        \setlength{\tabcolsep}{.06em}
        \begin{tabular}{cccccccc}
           \addlinespace[4pt]
            \includegraphics[width=0.1\textwidth]{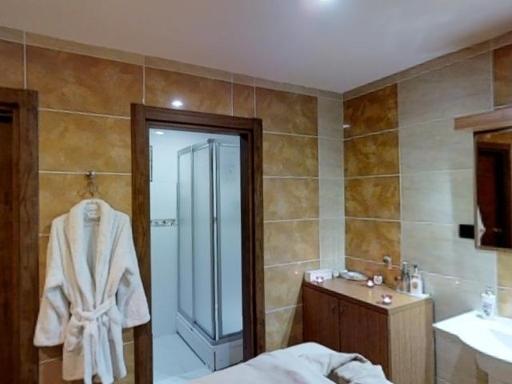} &
            \includegraphics[width=0.1\textwidth]{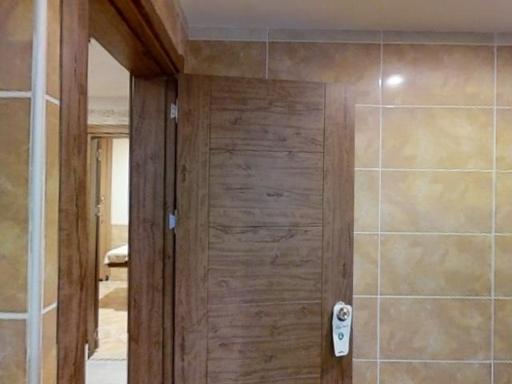} &
            \includegraphics[width=0.1\textwidth]{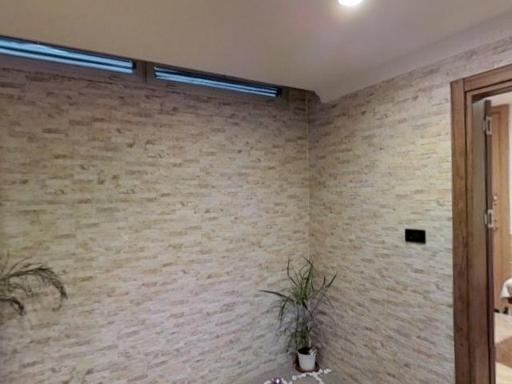} &
            \includegraphics[width=0.1\textwidth]{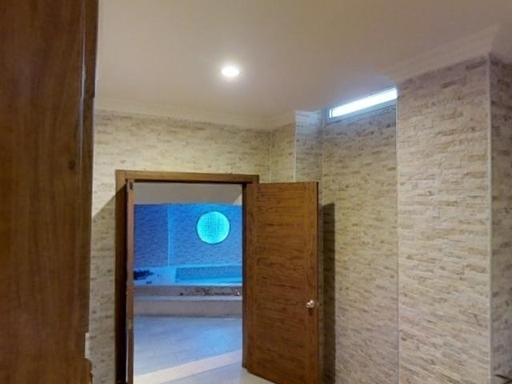} &
            \includegraphics[width=0.1\textwidth]{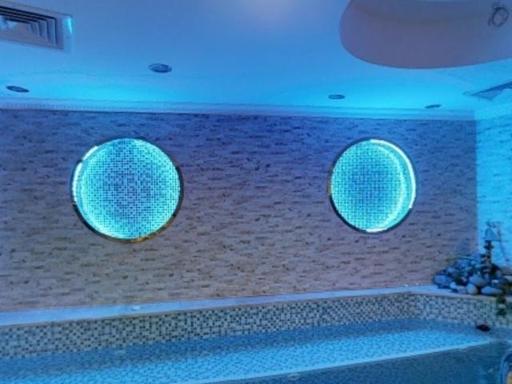} &
            \includegraphics[width=0.1\textwidth]{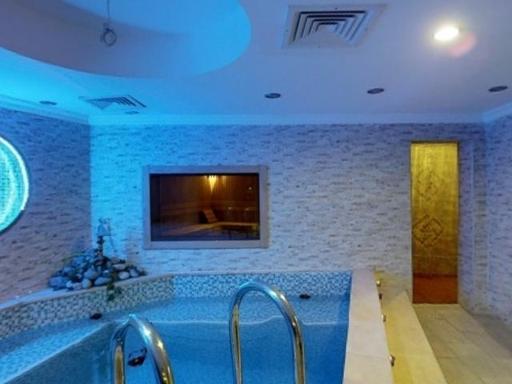} &
            \includegraphics[width=0.1\textwidth]{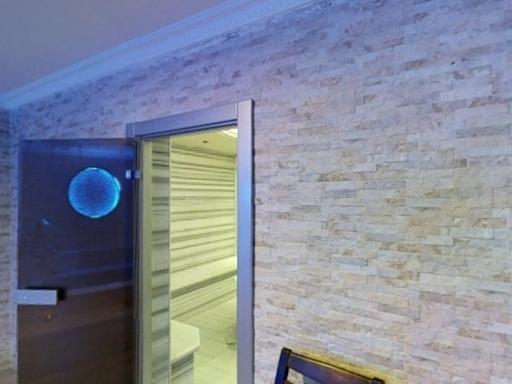} &
            \includegraphics[width=0.1\textwidth]{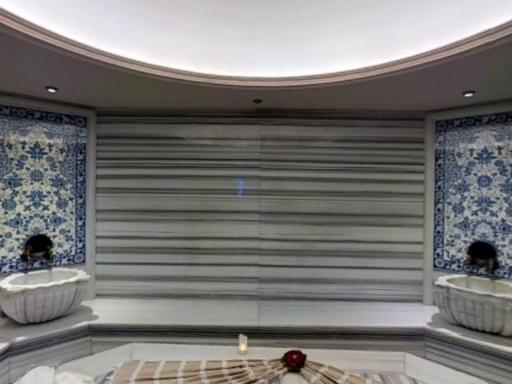} \\
        \end{tabular}
    }
    \resizebox{\linewidth}{!}{
    \centering
        \begin{tabular}{rp{160mm}}
           \multirow{2}{2cm}{\raggedleft (a)} & \textbf{GT}: \textit{Go to the white room with 4 basins opposite the lounge pool and turn on the faucet on the left closest to the entrance}
           \newline
           \textbf{AIGeN}: \textit{Walk to the bathroom with the white sink and turn on the faucet} \\
           \addlinespace[7pt]
        \end{tabular}
    }
    \resizebox{.99\linewidth}{!}{
        \setlength{\tabcolsep}{.06em}
        \begin{tabular}{cccccccc}
            \includegraphics[width=0.1\textwidth]{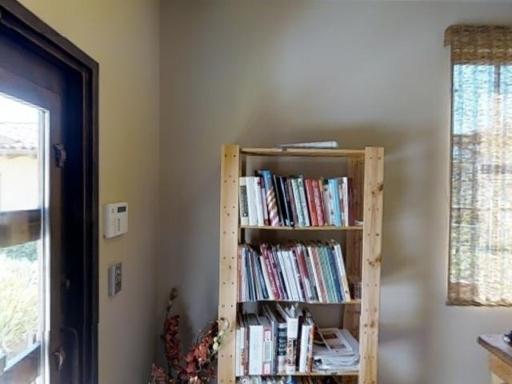} &
            \includegraphics[width=0.1\textwidth]{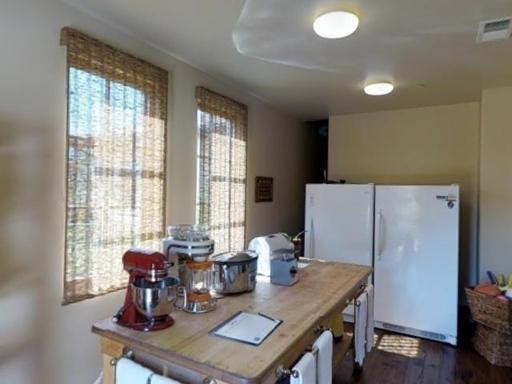} &
            \includegraphics[width=0.1\textwidth]{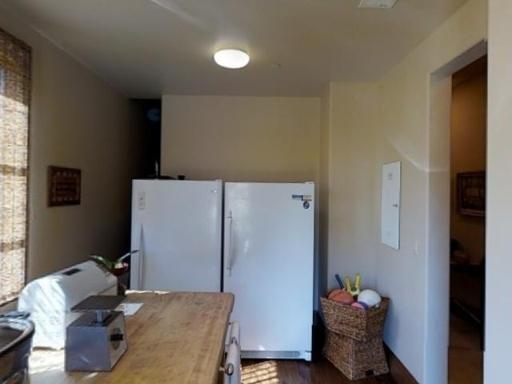} &
            \includegraphics[width=0.1\textwidth]{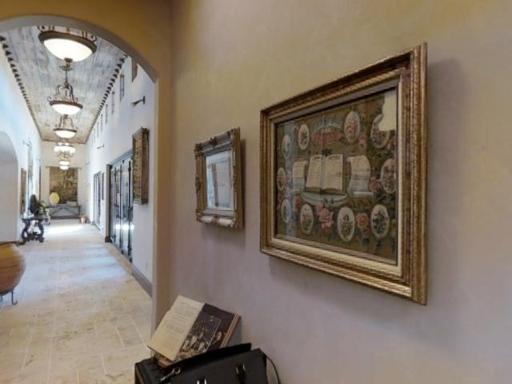} &
            \includegraphics[width=0.1\textwidth]{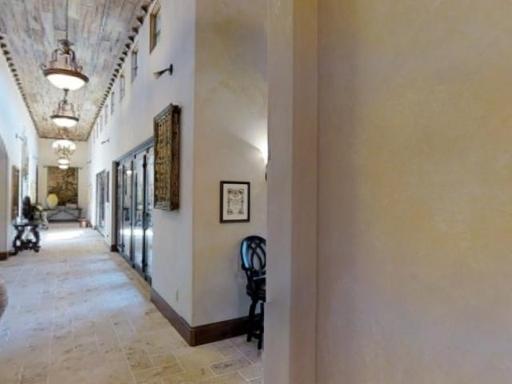} &
            \includegraphics[width=0.1\textwidth]{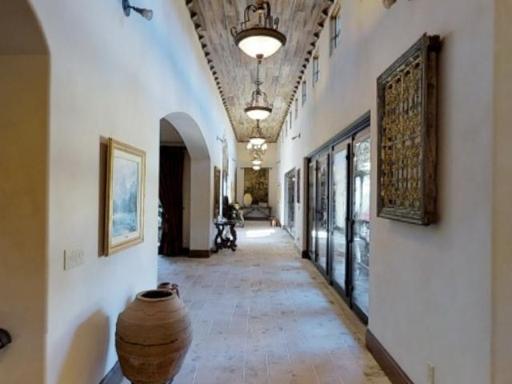} &
            \includegraphics[width=0.1\textwidth]{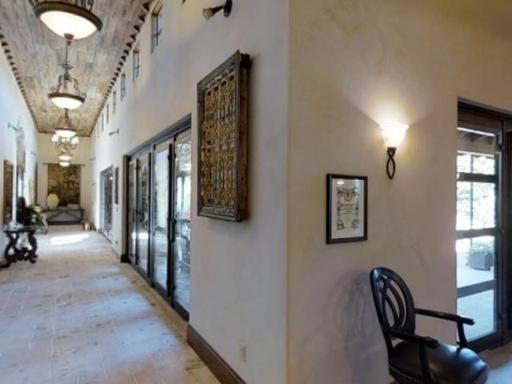} &
            \includegraphics[width=0.1\textwidth]{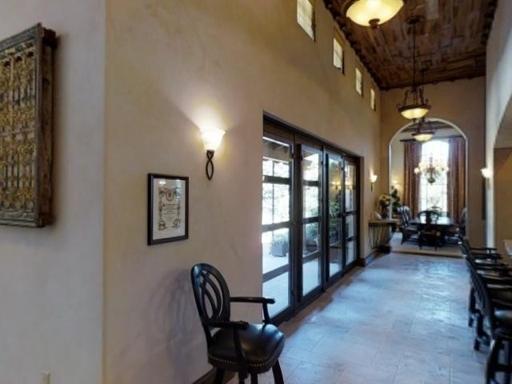} \\
        \end{tabular}
    }
    \resizebox{\linewidth}{!}{
    \centering
        \begin{tabular}{rp{160mm}}
           \multirow{2}{2cm}{\raggedleft (b)} & \textbf{GT}: \textit{Go to the hallway with the black chairs on level 2 and sit in the second chair from your right}
           \newline
           \textbf{AIGeN}: \textit{Go to the dining room and bring me the chair that is nearest to the door} \\
          \addlinespace[7pt]
        \end{tabular}
    }
    \resizebox{.99\linewidth}{!}{
        \setlength{\tabcolsep}{.06em}
        \begin{tabular}{cccccccc}
            \includegraphics[width=0.1\textwidth]{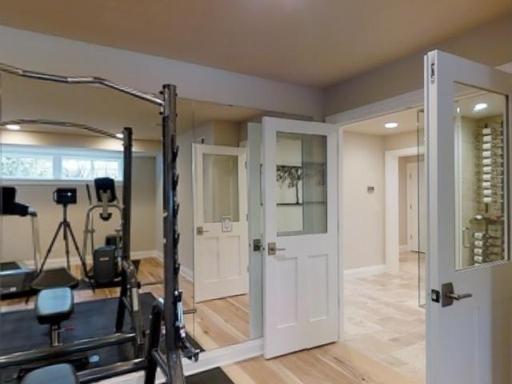} &
            \includegraphics[width=0.1\textwidth]{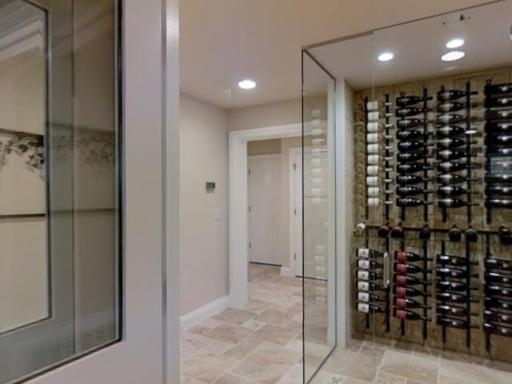} &
            \includegraphics[width=0.1\textwidth]{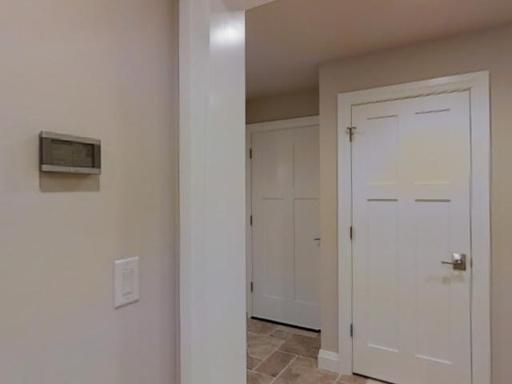} &
            \includegraphics[width=0.1\textwidth]{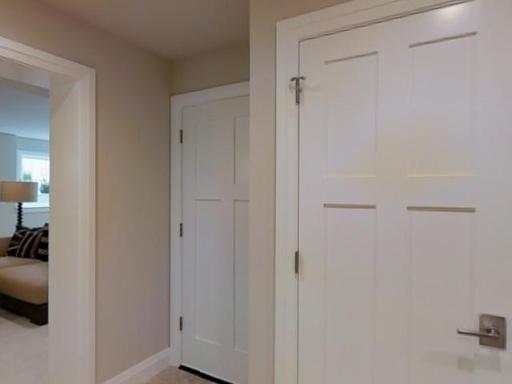} &
            \includegraphics[width=0.1\textwidth]{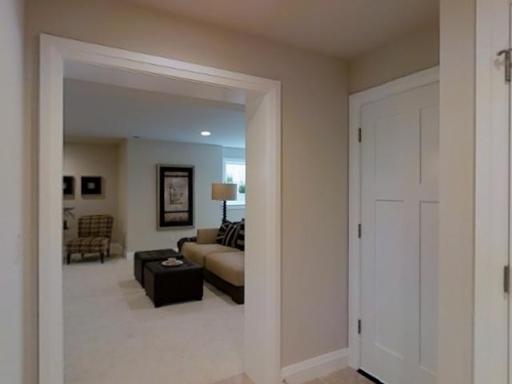} &
            \includegraphics[width=0.1\textwidth]{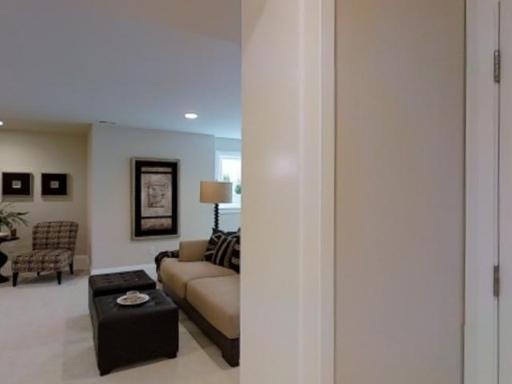} &
            \includegraphics[width=0.1\textwidth]{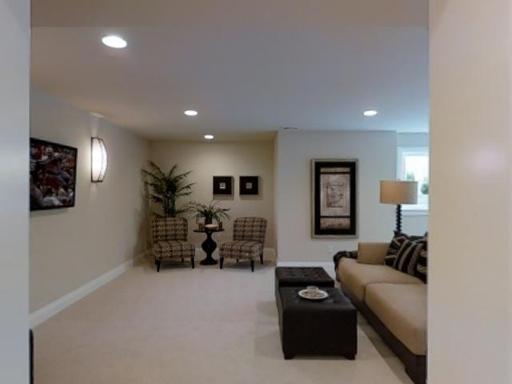} &
            \includegraphics[width=0.1\textwidth]{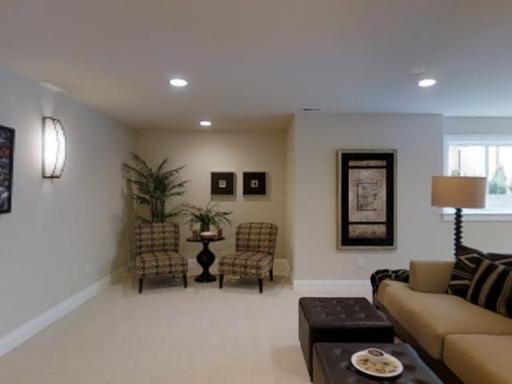} \\
        \end{tabular}
    }
    \resizebox{\linewidth}{!}{
    \centering
        \begin{tabular}{rp{160mm}}
           \multirow{2}{2cm}{\raggedleft (c)} & \textbf{GT}: \textit{Go to the lounge on level 1 and sit down on the armchair}
           \newline
           \textbf{AIGeN}: \textit{Go to the dining room and with a lint roller clean the arm chair which is against the wall} \\
           \addlinespace[7pt]
        \end{tabular}
    }
    \resizebox{.99\linewidth}{!}{
        \setlength{\tabcolsep}{.06em}
        \begin{tabular}{cccccccc}
            \includegraphics[width=0.1\textwidth]{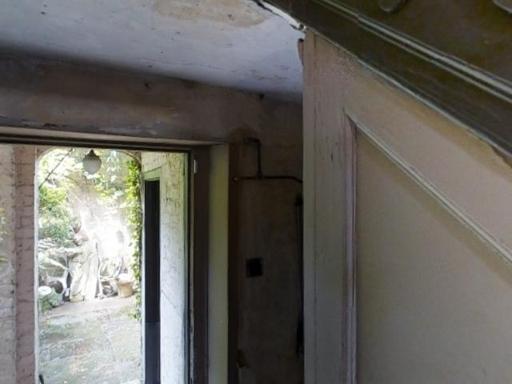} &
            \includegraphics[width=0.1\textwidth]{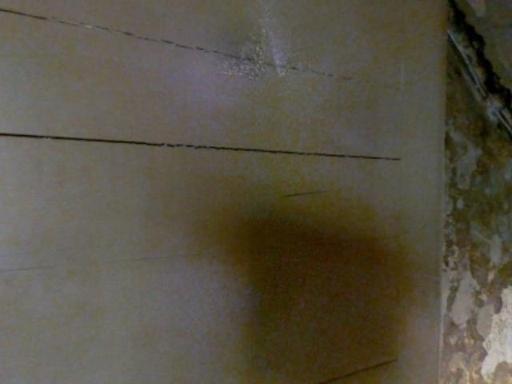} &
            \includegraphics[width=0.1\textwidth]{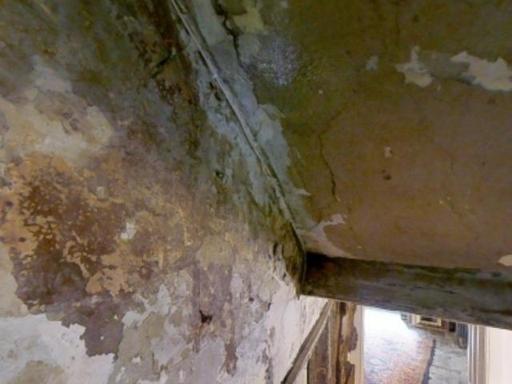} &
            \includegraphics[width=0.1\textwidth]{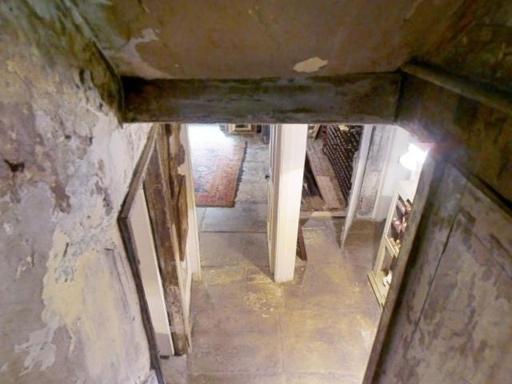} &
            \includegraphics[width=0.1\textwidth]{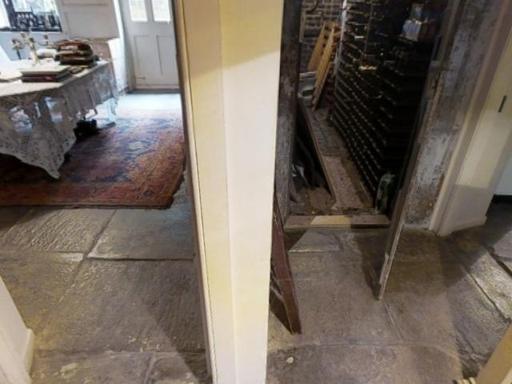} &
            \includegraphics[width=0.1\textwidth]{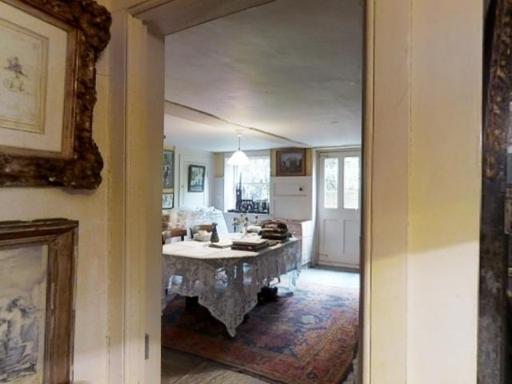} &
            \includegraphics[width=0.1\textwidth]{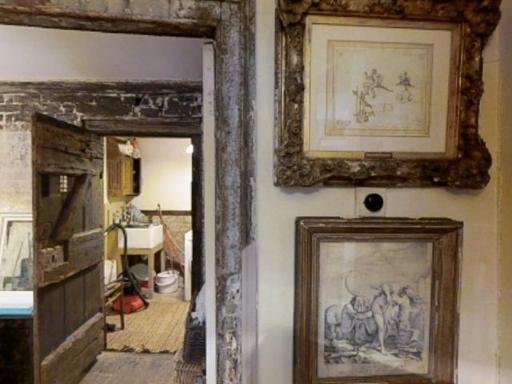} &
            \includegraphics[width=0.1\textwidth]{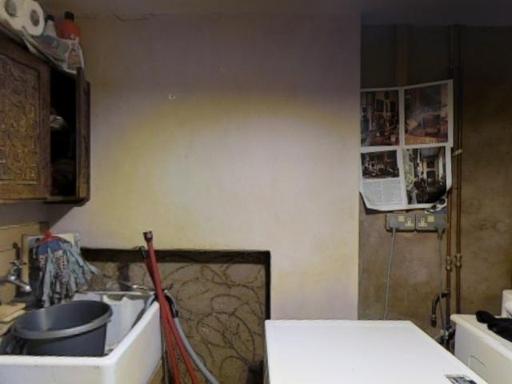} \\
        \end{tabular}
    }
    \resizebox{\linewidth}{!}{
    \centering
        \begin{tabular}{rp{160mm}}
           \multirow{2}{2cm}{\raggedleft (d)} & \textbf{GT}: \textit{Go to the laundry room on the first floor and sort out the cluttered cleaning products on the dryer}
           \newline
           \textbf{AIGeN}: \textit{Go to the laundry room on level 1 and bring me the clothes on the rack opposite the dryer} \\
          \addlinespace[7pt]
        \end{tabular}
    }
    \vspace{-2mm}
    \caption{Sample image sequences from REVERIE Val Unseen split with corresponding ground-truth instruction and synthetic instructions generated using AIGeN. The images in each sequence have been reduced to $8$ to facilitate the graphical presentation and we only show the frontal image of the panoramic observation at each timestep.}
    \label{fig:qualitative}
\end{figure*}

\subsection{Diversity Analysis}
\label{ssec:diversity}
To assess the quality of the synthetic instructions and measure the diversity of the generation, in Table~\ref{tab:diversity} we compute diversity metrics commonly used for image captioning~\cite{stefanini2022show,wang2020diversity} comparing the diversities of AIGeN-generated instructions with the ground-truth annotations of the trajectories of REVERIE training split. The comparison is performed by producing synthetic instruction using AIGeN before and after the adversarial fine-tuning phase.

The metrics used for this study are the ratio of the number of novel sentences, \ie~not contained in the dataset, to the number of ground-truth sentences, the number of unique words (Unigrams); the number of unique couples of consecutive words (Bigrams); Div-1; and Div-2. Div-1 and Div-2 are respectively the ratio of unique unigrams and bigrams to the total number of unigrams. 

Looking at the results, it is evident that the fine-tuning phase using an adversarial approach helps to generate instructions that do not retrace the ground truth instruction, with the consequence of improving the diversity of the dataset. In fact, while the generated instructions without using the GAN-like training present a small number of novel sentences, AIGeN returns a completely novel set of instructions. 
Furthermore, AIGeN with the adversarial fine-tuning can increase the number of unigrams and bigrams sampled from the word dictionary even with respect to the ground truth annotations.
This result is also reflected by Div-1 and Div-2 metrics that present a significant increase with respect to the ground truth instructions. 

\begin{figure*}[!ht]
    \resizebox{.99\linewidth}{!}{
        \setlength{\tabcolsep}{.06em}
        \begin{tabular}{cccccccc}
            \includegraphics[width=0.1\textwidth]{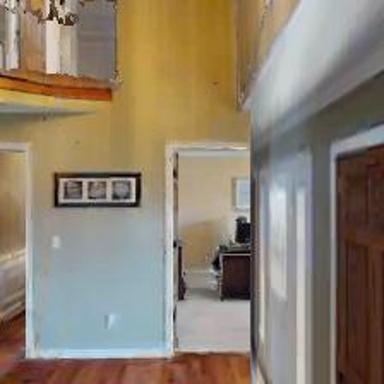} &
            \includegraphics[width=0.1\textwidth]{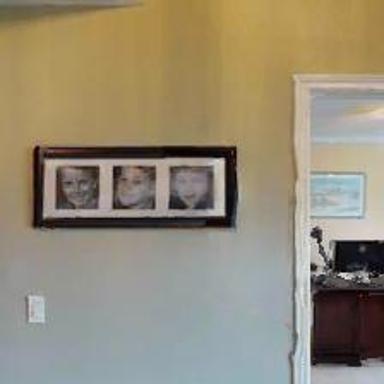} &
            \includegraphics[width=0.1\textwidth]{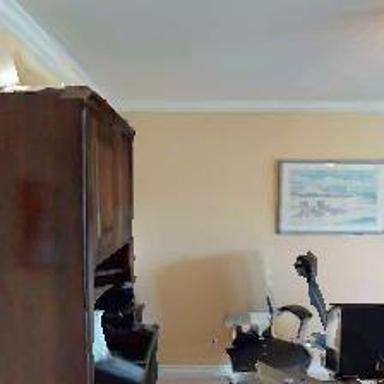} &
            \includegraphics[width=0.1\textwidth]{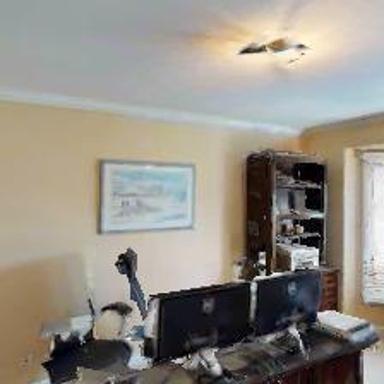} &
            \includegraphics[width=0.1\textwidth]{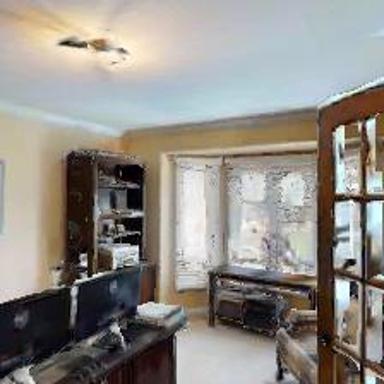} &
            \includegraphics[width=0.1\textwidth]{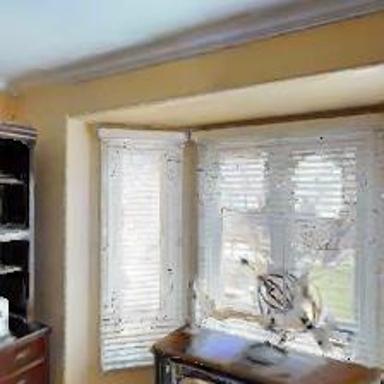} &
            \includegraphics[width=0.1\textwidth]{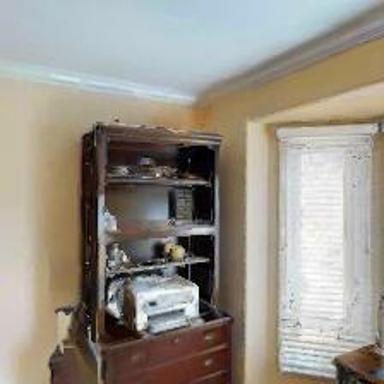} &
            \includegraphics[width=0.1\textwidth]{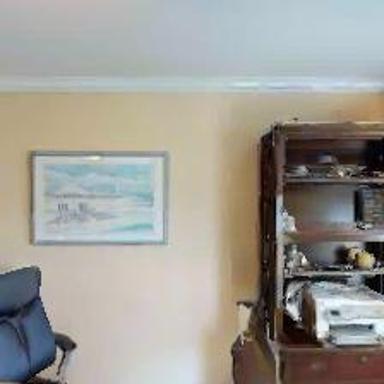} \\
        \end{tabular}
    }
    \centering
    \resizebox{\linewidth}{!}{
    \centering
        \begin{tabular}{rp{160mm}}
           \hspace{3cm}\raggedleft (e) &
           \textbf{AIGeN}: \textit{Go to the office on level 1 and clean the desk} \\
          \addlinespace[7pt]
        \end{tabular}
    }
    \resizebox{.99\linewidth}{!}{
        \setlength{\tabcolsep}{.06em}
        \begin{tabular}{cccccccc}
            \includegraphics[width=0.1\textwidth]{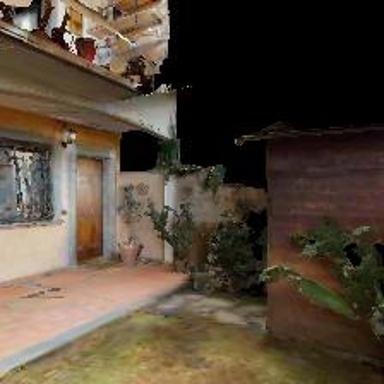} &
            \includegraphics[width=0.1\textwidth]{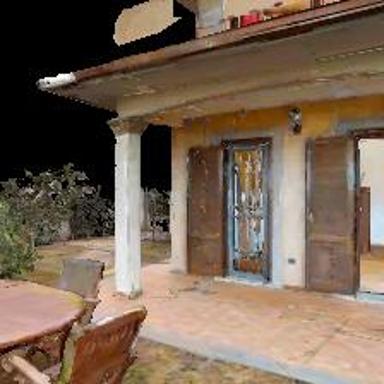} &
            \includegraphics[width=0.1\textwidth]{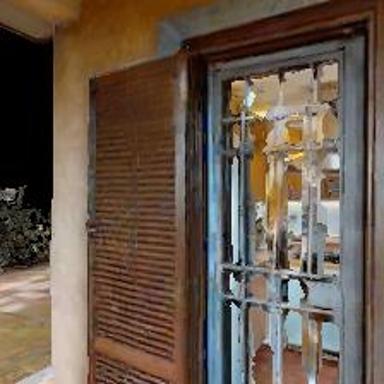} &
            \includegraphics[width=0.1\textwidth]{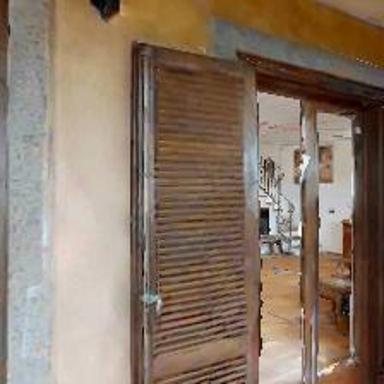} &
            \includegraphics[width=0.1\textwidth]{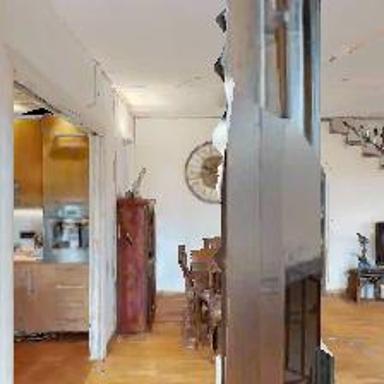} &
            \includegraphics[width=0.1\textwidth]{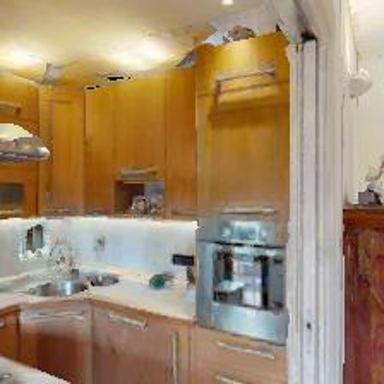} &
            \includegraphics[width=0.1\textwidth]{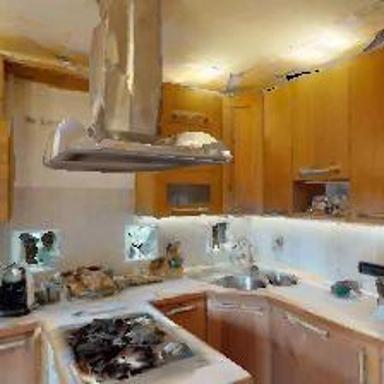} &
            \includegraphics[width=0.1\textwidth]{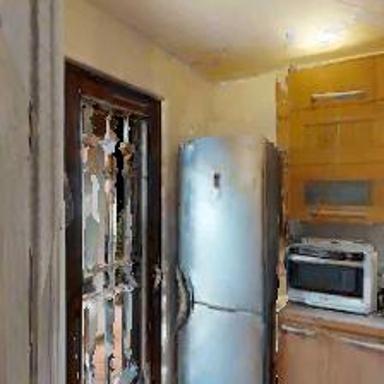} \\
        \end{tabular}
    }
    \centering
    \resizebox{\linewidth}{!}{
    \centering
        \begin{tabular}{rp{160mm}}
           \hspace{3cm}\raggedleft (f) & 
           \textbf{AIGeN}: \textit{Go to the kitchen and turn off the light closest to the entrance} \\
         \addlinespace[7pt]
        \end{tabular}
    }
    \resizebox{.99\linewidth}{!}{
        \setlength{\tabcolsep}{.06em}
        \begin{tabular}{cccccccc}
            \includegraphics[width=0.1\textwidth]{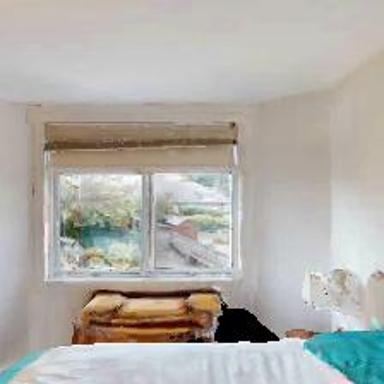} &
            \includegraphics[width=0.1\textwidth]{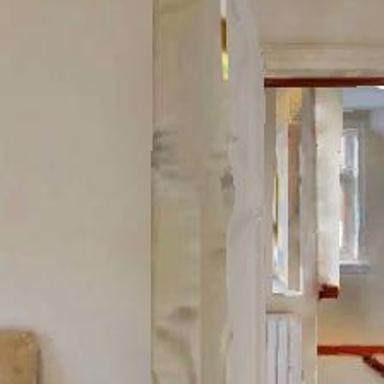} &
            \includegraphics[width=0.1\textwidth]{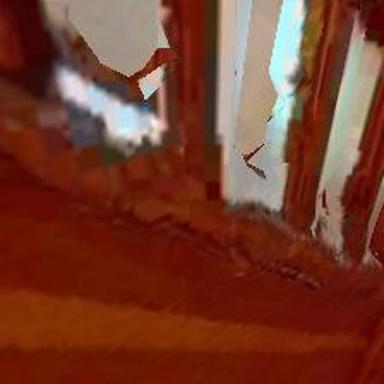} &
            \includegraphics[width=0.1\textwidth]{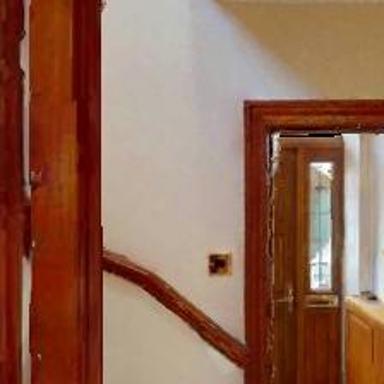} &
            \includegraphics[width=0.1\textwidth]{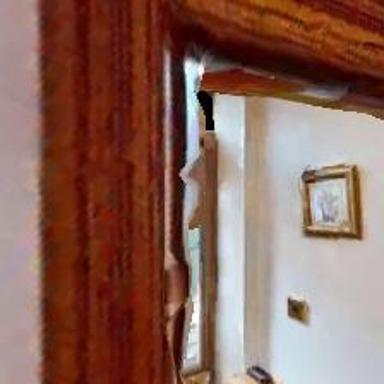} &
            \includegraphics[width=0.1\textwidth]{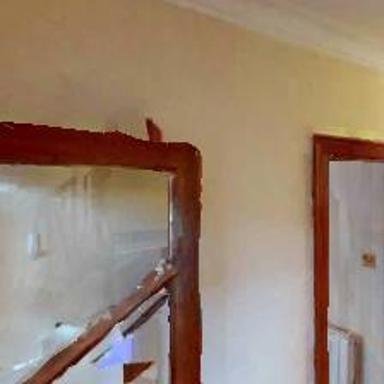} &
            \includegraphics[width=0.1\textwidth]{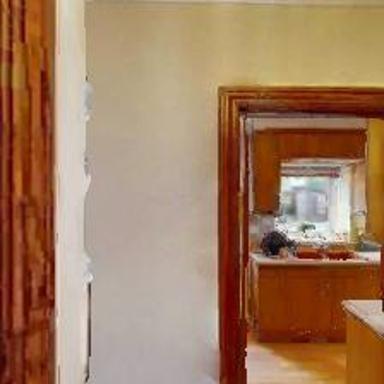} &
            \includegraphics[width=0.1\textwidth]{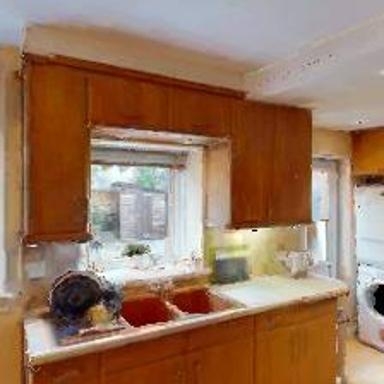} \\
        \end{tabular}
    }
    \centering
    \resizebox{\linewidth}{!}{
    \centering
        \begin{tabular}{rp{160mm}}
           \hspace{3cm}\raggedleft (g) &
           \textbf{AIGeN}: \textit{Go to the laundry room and clean the mirror above the sink} \\
         \addlinespace[7pt]
        \end{tabular}
    }
    \vspace{-4mm}
    \caption{Sample image sequences from HM3D with corresponding synthetic instructions generated using AIGeN. The images in each sequence have been reduced to $8$ to facilitate graphical presentation, and we only show the frontal image of the panoramic observation at each timestep.}
    \label{fig:qualitative_hm3d}
\end{figure*}
\subsection{Qualitative Analysis}
\label{ssec:qualitative}
Finally, Fig.~\ref{fig:qualitative} and~\ref{fig:qualitative_hm3d} show examples of sequences of input images with corresponding ground-truth instructions (if available) and generated instructions using AIGeN. 

In Fig.~\ref{fig:qualitative} all four samples have been taken from the ``unseen'' validation split of REVERIE, so that AIGeN has never seen these environments during training. We provide two positive samples in (a) and (b) as well as two negative ones in (c) and (d). For both (a) and (b), the generated instruction is similar to the ground-truth instruction and matches the given sequence of images. In fact, the landmarks bathroom in (a) and the dining room in the hallway in (b) are correctly recognized. Furthermore, both synthetic instructions refer to the correct target object, respectively ``faucet'' and ``chair''. In the case of (c) instead, the generated instruction and the ground-truth instruction identify the wrong landmark, with the lounge that is recognized as a dining room by our model. 
However, the target object ``armchair'' is still recognized correctly. 
Finally, in the case of (d), the generated instruction and the ground-truth one, refer to different target objects. While the correct target objects are the ``cluttered cleaning products'' on the dryer, AIGeN refers to the ``clothes on the rack'' as the target objects. Nevertheless, the laundry room is correctly identified by our model. 

Moving on to Fig.~\ref{fig:qualitative_hm3d}, we showcase three qualitative samples using trajectories from HM3D. As HM3D is unlabeled, in this case, there is no ground-truth annotation available, and we only provide the instructions generated by AIGeN.
The first example in (e) identifies the office and the target object ``desk'' correctly in the final observation of the sequence. 
In the second trajectory shown in (f), the target object (``light'') in the kitchen is correctly recognized.
Finally, in the third trajectory (g), the kitchen is identified as a laundry room, and the correct target object ``window'' is misidentified as a ``mirror''.

These results demonstrate that even when the visual quality of the environment is low due to 3D reconstruction, AIGeN is capable of generating valuable instructions for vision-and-language navigation providing correct directions identifying objects and landmarks.
\section{Conclusions}
\label{sec:conclusion}
We proposed a computational model for synthetic instruction generation, based on the state-of-the-art NLP models GPT-2 and BERT. Our model leverages a GAN-like structure and takes sequences of images as input to generate instructions for the path traversed by the agent to reach its target location.
The model has been trained and validated on REVERIE instructions and achieved high image description metric results
when comparing the generated instructions with the ground-truth instructions.
Subsequently, we prove that using our synthetic instructions to augment a VLN dataset, such as REVERIE or R2R, improves the performance of a VLN method achieving state-of-the-art performance on both navigation and object grounding metrics.
\section{Limitations}
\label{sec:limitations}
In our experimental section, we extract the labels of the objects in the last panoramic view of each navigation trajectory using Mask2Former \cite{cheng2022masked} trained on ADE20K. The labels of these objects are tokenized to obtain their embeddings which are then concatenated with the image embeddings and the instructions embeddings. 
Relying on detection labels the model can identify the target objects within each episode and potentially understand the surrounding context and landmarks. This time-consuming procedure is necessary as the words derived from the object labels help to generate coherent instructions for the provided image sets.

As can be seen in the ablation study presented in Table~\ref{table:description_ablation} and Table~\ref{table:navigation_ablation}, the instruction quality and navigation performance of AIGeN without the object detections are worse than those of the complete approach. 
Future work could focus on generating instructions directly from the image sequence trying to fill the gap with the approach using upstream object detection.
\section*{Acknowledgments}
Niyati Rawal and Roberto Bigazzi were supported by Marie Sklodowska-Curie Action Horizon 2020
(Grant agreement No. 955778) for the project ``Personalized Robotics as Service Oriented Applications'' (``PERSEO''). Lorenzo Baraldi and Rita Cucchiara were supported by the ``Fit for Medical Robotics'' (``Fit4MedRob'') project, funded by the Italian Ministry of University
and Research.
{
    \small
    \bibliographystyle{ieeenat_fullname}
    \bibliography{egbib}
}


\end{document}